\documentclass[noindent]{article}
\usepackage{amsmath}
\usepackage{amssymb}
\usepackage{theorem}
\newtheorem{defn}{Definition}
\newtheorem{thm}{Theorem}

\begin{document}
\title{Multilayer Perceptron Algebra \\-- A Mathematical Theory on the Design of Neural Networks}
\author{Zhao Peng \\ zhaopengmath@mail.bnu.edu.cn}
\maketitle

\begin{abstract}
    Artificial Neural Networks(ANN) has been phenomenally successful on various pattern recognition tasks. However, 
    the design of neural networks rely heavily on the experience and intuitions of individual developers.
    In this article, the author introduces a mathematical structure called MLP algebra on the set of all Multilayer Perceptron Neural Networks(MLP), which can serve as a guiding principle to build MLPs accommodating to the particular data sets, and to build complex MLPs from simpler ones.\\\\
\end{abstract}

\section{Introduction}

Consider this simple question:

$\mathbb{D}_1, \mathbb{D}_2$ are two different data sets with the same label, while $\mathcal{N}_1, \mathcal{N}_2$ are two neural networks achieving satisfying classification
results on these two data sets respectively. 
Let $\mathbb{D} = \mathbb{D}_1 \cup \mathbb{D}_2$. 
How to design a new neural network $\mathcal{N}$ to achieve satisfying classification results on $\mathbb{D}$?\\

It is reasonable to guess that the new network is related to $\mathcal{N}_1, \mathcal{N}_2$.
In fact, they are related in a very elaborate, mathematically explicit way. 
This article will reveal this relationship, while establishing a whole family of connections of the same kind.

The main idea is that complex neural networks can be built from simpler ones. A neural network aiming at classifying a complex dataset can be very hard to train, but it is usually 
possible to decompose a complex dataset into certain combinations of simpler sets, and
simpler sets are easier for a neural network to recognize. 
If we have an explicit algorithm to 
integrate the networks trained by these simpler sets into a new network while maintaining 
the recognition accuracy on each of the sets, we manage to build a network that works well 
on the original set.

To carry out such a construction process, the author introduces the MLP-algebra, which is a
set of operations that enable us to do arithmetic on MLP neural networks, such as adding 
two networks, substract a network from another, multiply two neural networks, and decompose a network into simpler components.
These operations proves to be very handy to build neural networks when certain information of the dataset is given.

\section{Notations}

In case of undesired ambiguity, let us agree on the following definitions of terms.

\subsection{On Data Set}
\begin{enumerate}
\item A \textbf{Labeled Data Set} is a couple of matrices $(\mathbb{D}, \mathbb{L})$. 

\item The \textbf{Data Set} $\mathbb{D}$ is a set of $m$ vectors in $\mathbb{R}^n$. It can be arranged as an $m \times n$ matrix, with $m$ representing the number of data points,
and $n$ representing the dimension of the underlying space(a Euclidean space).

\item The \textbf{Label Set} $\mathbb{L}$ is a $m \times 1$ matrix, which contains the vectorized labels in corespondens with $\mathbb{D}$.

\item $\mathbb{D}_{\epsilon} = \cup_{p \in \mathbb{D}} B(p, \epsilon)$ is called the $\epsilon$--neighbourhood of $\mathbb{D}$. $B(p, \epsilon)$ is the ball centered at $p$ with radius $\epsilon$.
\end{enumerate}

The MNIST database (Mixed National Institute of Standards and Technology) is a typical dataset.
It contains 70000 images (60000 for training and 10000 for testing) of hand written digits , each casted to $ 28 \times 28$ pixels, with a label from 0 to 9.

\subsection{On Network}
\begin{enumerate}
\item An \textbf{Multi-Layer Perceptron Neural Network(MLP)} $\mathcal{N}$ is a mapping between two Euclidean spaces $(\mathbb{R}^{n_1}, \mathbb{R}^{n_L}).$
    The mapping is defined by a sequence of Euclidean spaces 
    \[(\mathbb{R}^{n_1}, \mathbb{R}^{n_2}, \dots, \mathbb{R}^{n_L})\]
    and the mappings connecting them:
    \[f_1: \mathbb{R}^{n_1} \rightarrow \mathbb{R}^{n_2}, f_2: \mathbb{R}^{n_2} \rightarrow \mathbb{R}^{n_3}, \dots, f_{L-1}: \mathbb{R}^{n_{L-1}} \rightarrow \mathbb{R}^{n_L}\]
\item The connecting functions are usually given by the sigmoid function $\sigma$.
    \[f_{i,j}(x) = \sigma(\sum_{k = 1}^{n_i}\omega^i_{jk}x_k - \theta^i_j), \]
    \[\sigma(z) = \frac{1}{1 + e^{-z}}\]
\item Here the $\omega\,$s are called \textbf{weights}, $\omega^i$ denotes the \textbf{weight matrix} ($n_{i+1} \times n_i$) connecting layer $i$ and $i+1$.
\item The $\theta\,$s are called \textbf{thresholds}. $\theta^i$ denotes the \textbf{threshold matrix} ($n_{i+1} \times 1$) connecting layer $i$ and $i+1$.
    \item We denote by $b^k(\mathcal{N})$ the \textbf{k-th layer} (an Euclidean space) of the network. For example, $b^1(\mathcal{N}) = \mathbb{R}^{n_1}$ for the network above.
\end{enumerate}

Neural Networks are used to classify datasets. If the dataset to be classified has only 1 category (label), 
it simply asks for a yes-or-no answer, a network with a 1-dimensional output layer will suffice. 
The output value of this single neuron simply means the probability of a data point belonging to this category.
When there is multiple categories (labels), as in the case of the MNIST database, a network
with a corresponding output layer dimension will come very handy. 
More specifically, the original label needs to be vectorized. 
For exmaple, the label 0 in the MNIST database has to be converted to a 10-dimensional vector $(1, 0, \dots, 0)$, label 1 converted to the 10-diemnsional vector $(0, 1, 0, \dots, 0)$ , \emph{etc}.
The output value of the network is also a 10-dimensional vector, and the largest entry is our classification result.

\section{MLP Algebra}

\subsection{Network Operations}

\begin{defn}[Complementary Net]\label{complement}
    The \textbf{Complementary Net} ,$\mathcal{N}^c$, of an L-layered MLP $\mathcal{N}\,(L \geq 3)$ is an L-layered MLP defined by the following equations:
    \begin{enumerate}
        \item On layer structure:
            \begin{align*}
                &b^k(\mathcal{N}^c) = b^k(\mathcal{N}), \\
            \end{align*}
        \item On weights:
            \begin{align*}
            &\omega^k(\mathcal{N}^c) = \omega^k(\mathcal{N}), \,\, k = 1, 2, \dots, L-2\\
            &\omega^{L-1}(\mathcal{N}^c) = -\omega^{L-1}(\mathcal{N}^c), \\
            \end{align*}
        \item On thresholds:
            \begin{align*}
                &\theta^k(\mathcal{N}^c) = \theta^k(\mathcal{N}), \,\, k = 1, 2, \dots, L-2\\
                &\theta^{L-1}(\mathcal{N}^c) = -\theta^{L-1}(\mathcal{N}^c)
             \end{align*}  
 \end{enumerate}
\end{defn}
It is easily verified that $\mathcal{N}(x) = 0 \Rightarrow \mathcal{N}^c(x) = 1,\,\mathcal{N}(x) = 1 \Rightarrow \mathcal{N}^c(x) = 0$, which justifies the definition.

\begin{defn}[Sum Net]\label{sum}
    Let $\mathcal{N}_1, \mathcal{N}_2$ be two L-layered MLPs, with $b^1(\mathcal{N}_1) = b^1(\mathcal{N}_2)$ and $b^L(\mathcal{N}_1) = b^L(\mathcal{N}_2) = \mathbb{R}$. Their \textbf{Sum Net}, $\mathcal{N}_1 + \mathcal{N}_2$, is defined by the following equations:
    \begin{enumerate}
        \item On layer structure:
            \begin{align*}
            &b^1(\mathcal{N}_1 + \mathcal{N}_2) = b^1(\mathcal{N}_1), \\
            &b^1(\mathcal{N}_1 + \mathcal{N}_2) = b^1(\mathcal{N}_2) \\
            &b^k(\mathcal{N}_1 + \mathcal{N}_2) = b^k(\mathcal{N}_1)\times b^k(\mathcal{N}_2), \,
             k = 2, \dots, L\\
             &b^{L+1} = \mathbb{R}
            \end{align*}
        Here the times symbol $\times$ means Descartes product between sets.
        \item On weights:
            \begin{align*}
            &\omega^1(\mathcal{N}_1 + \mathcal{N}_2) = 
             \left(\begin{array}{c}\omega^1(\mathcal{N}_1)\\\omega^1(\mathcal{N}_2)\end{array}\right)\\
            &\omega^k(\mathcal{N}_1 + \mathcal{N}_2) = 
                 \left(\begin{array}{cc}
                         \omega^k(\mathcal{N}_1)&\\
                                                &\omega^k(\mathcal{N}_2)
                     \end{array}\right),\,
                     &k = 2, \dots, L-1\\
                     &\omega^{L}(\mathcal{N}_1 + \mathcal{N}_2) = \lambda(1, 1)\\
             \end{align*}
         \item On thresholds:
             \begin{align*}
            &\theta^k(\mathcal{N}_1 + \mathcal{N}_2) = 
                 \left(\begin{array}{c}
                     \theta^k(\mathcal{N}_1)\\
                     \theta^k(\mathcal{N}_2) 
             \end{array}\right)
             &k = 1, 2, \dots, L-1\\
             &\theta^L(\mathcal{N}_1 + \mathcal{N}_2) = 0.5\lambda
            \end{align*}
        \end{enumerate}
\end{defn}

However, this operation cannot be applied directly to multi-variable case, for the sum net has
one more layer. A generalised definition can help to avoid this problem.

\begin{defn}[Multi-Sum Net]
    Let $\mathcal{N}_1, \mathcal{N}_2, \dots, \mathcal{N}_m$ be $m$ L-layered MLPs, 
    with $b^1(\mathcal{N}_1) = b^1(\mathcal{N}_2) = \cdots = b^1(\mathcal{N}_m)$ 
    and $b^L(\mathcal{N}_1) = b^L(\mathcal{N}_2) = \cdots = b^L(\mathcal{N}_m) = \mathbb{R}$. 
    Their \textbf{Multi-Sum Net}, $\sum_{i=1}^k\mathcal{N}_i$, is defined by the following equations:
    \begin{enumerate}
        \item On layer structure:
            \begin{align*}
                &b^1(\sum_{i=1}^k\mathcal{N}_i) = b^1(\mathcal{N}_j), &j = 1, 2, \dots, m\\
            &b^k(\sum_{i=1}^k\mathcal{N}_i) = b^k(\mathcal{N}_1)\times 
                                              b^k(\mathcal{N}_2)\times
                                              \cdots\times
                                              b^k(\mathcal{N}_m), \,
                                              &k = 2, \dots, L\\
             &b^{L+1}(\sum_{i=1}^k\mathcal{N}_i) = \mathbb{R}
            \end{align*}
        \item On weights:
            \begin{align*}
            &\omega^1(\sum_{i=1}^k\mathcal{N}_i) = 
             \left(\begin{array}{c}
             \omega^1(\mathcal{N}_1)\\
             \omega^1(\mathcal{N}_2)
             \end{array}\right)\\
            &\omega^k(\sum_{i=1}^k\mathcal{N}_i) = 
                 \left(\begin{array}{cc}
                         \omega^k(\mathcal{N}_1)&\\
                                                &\omega^k(\mathcal{N}_2)
                     \end{array}\right)
                     &k = 2, \dots, L-1\\
                     &\omega^{L}(\sum_{i=1}^k\mathcal{N}_i) = \lambda(1, 1, \dots, 1)\\
             \end{align*}
         \item On thresholds:
             \begin{align*}
            &\theta^k(\sum_{i=1}^k\mathcal{N}_i) = 
                 \left(\begin{array}{c}
                     \theta^k(\mathcal{N}_1)\\
                     \theta^k(\mathcal{N}_2) 
             \end{array}\right)
             &k = 1, 2, \dots, L-1\\
             &\theta^L(\sum_{i=1}^k\mathcal{N}_i) = 0.5\lambda
            \end{align*}
        \end{enumerate}
\end{defn}

Combine definitions \ref{complement} and \ref{sum}, we come to

\begin{defn}[Difference Net]
    Let $\mathcal{N}_1, \mathcal{N}_2$ be two L-layered MLPs, with $b^1(\mathcal{N}_1) = b^1(\mathcal{N}_2)$ and $b^L(\mathcal{N}_1) = b^L(\mathcal{N}_2) = \mathbb{R}$. Define 
    \[\mathcal{N}_1 - \mathcal{N}_2 = \mathcal{N}_1 + \mathcal{N}_2^c\]
        as their \textbf{Difference Net}
\end{defn}

\begin{defn}[I-Product Net]
    Let $\mathcal{N}_1, \mathcal{N}_2$ be two L-layered MLPs with $b^L(\mathcal{N}_1) = b^L(\mathcal{N}_2) = \mathbb{R}$, their \textbf{I-Product Net},
    $\mathcal{N}_1 \times \mathcal{N}_2$, is defined by the following equations:
    \begin{enumerate}
            \item On layer structure:
        \begin{align*}
            &b^k(\mathcal{N}_1 \times \mathcal{N}_2) = (b^k(\mathcal{N}_1), b^k(\mathcal{N}_2)), k = 1, 2, \dots, L\\
        &b^{L+1}(\mathcal{N}_1 \times \mathcal{N}_2) = \mathbb{R}\\
        \end{align*}
    \item On weights:
        \begin{align*}
        &\omega^k(\mathcal{N}_1 \times \mathcal{N}_2) = diag(\omega^k(\mathcal{N}_1),\omega^k(\mathcal{N}_2))\\
        &\omega^L(\mathcal{N}_1 \times \mathcal{N}_2) = \lambda(1, 1)\\
        \end{align*}
    \item On thresholds:
        \begin{align*}
        &\theta^k(\mathcal{N}_1 \times \mathcal{N}_2) = 
        \left(\begin{array}{c}
                \theta^k(\mathcal{N}_1)\\
                \theta^k(\mathcal{N}_2))
        \end{array}\right)\\
        &\theta^L(\mathcal{N}_1 \times \mathcal{N}_2) = 1.5\lambda
        \end{align*}
    \end{enumerate}
\end{defn}

In resemblence to the sum net, we can generalize this definition to multi-variable case.

\begin{defn}[Multi-I-Product Net]
    Let $\mathcal{N}_1, \mathcal{N}_2, \dots, \mathcal{N}_m$ be $m$ L-layered MLPs with $b^L(\mathcal{N}_1) = b^L(\mathcal{N}_2) = \cdots = b^L(\mathcal{N}_m) = \mathbb{R}$, their \textbf{Multi-I-Product Net},
    $\prod_{i = 1}^m\mathcal{N}_i$, is defined by the following equations:
    \begin{enumerate}
            \item On layer structure:
        \begin{align*}
            &b^k(\prod_{i = 1}^m\mathcal{N}_i) = (b^k(\mathcal{N}_1), b^k(\mathcal{N}_2)), k = 1, 2, \dots, L\\
        &b^{L+1}(\prod_{i = 1}^m\mathcal{N}_i) = \mathbb{R}\\
        \end{align*}
    \item On weights:
        \begin{align*}
        &\omega^k(\prod_{i = 1}^m\mathcal{N}_i) = diag(\omega^k(\mathcal{N}_1),\omega^k(\mathcal{N}_2))\\
        &\omega^L(\prod_{i = 1}^m\mathcal{N}_i) = \lambda(1, 1)\\
        \end{align*}
    \item On thresholds:
        \begin{align*}
        &\theta^k(\prod_{i = 1}^m\mathcal{N}_i) = 
        \left(\begin{array}{c}
                \theta^k(\mathcal{N}_1)\\
                \theta^k(\mathcal{N}_2))
        \end{array}\right)\\
        &\theta^L(\prod_{i = 1}^m\mathcal{N}_i) = 1.5\lambda
        \end{align*}
    \end{enumerate}
\end{defn}

The networks in the definitions above all have 1-dimensional output layers, but the operations can be easily generalized, so long as the networks being summed (or multiplied) have the same output layer structure.

\subsection{Characteristic MLP}
\begin{defn}[Characteristic MLP]
    Let $\mathbb{D}$ be a data set. $\epsilon$ is a positive number.
    An MLP $\mathcal{N}_{\mathbb{D}_{\epsilon}}$ with 1-dimensional output layer is called an
    $\epsilon$ characteristic MLP of $\mathbb{D}$, if
    \begin{align*}
        &\mathcal{N}_{\mathbb{D}_{\epsilon}}(\mathbb{D}) = 1\\
        &\mathcal{N}_{\mathbb{D}_{\epsilon}}(\mathbb{D}_{\epsilon}^c) = 0
    \end{align*}
\end{defn}

When there is no danger of ambiguity, we denote $\mathcal{N}_{\mathbb{D}_{\epsilon}}$ by 
$\mathcal{N}_{\mathbb{D}}$ for short.\\

A perfect Characteristic MLP, of course, does not exit, as the output of a network with sigmoid linking is never strictly 0 or 1. 
But such networks can approximate the Characteristic MLPs quite well, so we intentionally blur their difference to maintain the simplicity of this theory.

It is very easy to train a characteristic MLP for datasets with simple geometry. 
For example, if $\mathbb{D} = \{x\,\in\mathbb{R}^n | \parallel x \parallel < 1\}$, 
an $n \times (n+1) \times 1$ MLP will converge very quickly to $\mathcal{N}_{\mathbb{D}}$ 
in a stochastic gradient descent process.

\begin{defn}[Accuracy of an MLP]\label{accuracy}
    Let $\mathcal{N}$ be an MLP, $\mathbb{D}$ is a labeled dataset, we define $\mathcal{A}_{\mathbb{D}}(\mathcal{N)}$ as its accuracy on $\mathbb{D}$:
\end{defn}

The definition above needs some further explaination.
\begin{enumerate}
\item If the output layer of $\mathcal{N}$ is 1-dimensional and $\mathbb{D}$ has only one label,
        then those $p \in \, \mathbb{D}$ with $\mathcal{N}(p) \geq 0.5$ are classified as belonging to the category.
\item If the output layer of $\mathcal{N}$ is $n_L$-dimensional and $\mathbb{D}$ has $n_L$ labels, 
    the $p \in \, \mathbb{D}$ is classified to the $argmax_{i \in \{1, \dots, n_L\}}\mathcal{N}(p)_i$-th label.
\end{enumerate}

The accuracy here simpy means number of correctly classified data points. 
When there is no danger of ambiguity, we take the liberty to omit the subscript for $\mathcal{A}$.

\begin{thm}[Basic Properties of Characteristic MLP]
\ 
    \begin{enumerate}
        \item If $\mathbb{D}$ is a data set, we have:
            \[\mathcal{A}(\mathcal{N}_{\mathbb{D}}^c) = \mathcal{A}(\mathcal{N}_{\mathbb{D}^c})\]
        \item If $\mathbb{D}_1, \mathbb{D}_2$ are two subsets of $\mathbb{R}^n$, we have:
            \[\mathcal{A}(\mathcal{N}_{\mathbb{D}_1 \cup \mathbb{D}_2}) =
            \mathcal{A}(\mathcal{N}_{\mathbb{D}_1} + \mathcal{N}_{\mathbb{D}_2})  \]
        \item If $\mathbb{D}_1 \subset \mathbb{R}^{n_1}, \mathbb{D}_{2} \subset \mathbb{R}^{n_2}$, we have:
            \[\mathcal{A}(\mathcal{N}_{\mathbb{D}_1 \times \mathbb{D}_2}) =
            \mathcal{A}(\mathcal{N}_{\mathbb{D}_1} \times \mathcal{N}_{\mathbb{D}_2}) \]
    \end{enumerate}
\end{thm}

The theorem is quite straight-forward, as all our operations in MLP algebra have very explicit logical implications.\\

A natural question is that, is this theorem still correct when we replace the Characteristic MLPs with general MLPs? 
As it turns out, the equations above will generally fail. but if we take the networks on the right hand side of those equations, train them for a bit more periods, the equations tend to hold again.
But if we create a new network with the same layer structure as the nets on the right hand side, the training process can be much harder.\\

The following short example will illustrate the potential of these simple properties. 
Suppose we want to build a characteristic MLP for a torus $\mathbb{T}^2$ in $\mathbb{R}^4$
(If we place the torus in $\mathbb{R}^3$ there will be some technical issues). 
The torus can be viewed as $\mathbb{S}^1 \times \mathbb{S}^1$, each $\mathbb{S}^1$ is a circle in $\mathbb{R}^2$. We have
\[\mathcal{N}_{\mathbb{T}^2} = \mathcal{N}_{\mathbb{S}^1} \times\mathcal{N}_{\mathbb{S}^1}\]

$\mathbb{S}^1$ can be viewed a large disk $\mathbb{D}_R$ minus a smaller one $\mathbb{D}_r$, hence
\[\mathcal{N}_{\mathbb{S}^1} = \mathcal{N}_{\mathbb{D}_R} - \mathcal{N}_{\mathbb{D}_r}\]
And as mentioned above, $\mathcal{N}_{\mathbb{D}}$ is very easy to train. We can see that the original, seemingly clueless task is ascribed to a composition of trivial ones.

\section{More MLP Algebra}

The network arithmetic discussed above are constrained to MLPs with 1-dimensional output layer, and the operands of sum, i-product and o-product are required to have the same layer structure, which is rare for pragmatic MLPs. 

Observed that any MLP can be decomposed into components. To make MLP algebra more practical, we introduce the following operations, mainly concerned with the combination and decomposition of MLPs.

\begin{defn}[Component Net]
    $\mathcal{N}$ is an L-layered MLP, $b^L(\mathcal{N}) = \mathbb{R}^{n_L}$, the \textbf{l-th Component Net}($ l \leq n_L$) $\,\, \mathcal{N}^{(l)}$is defined by the following equations:
    \begin{enumerate}
        \item On layer structure:
            \begin{align*}
            & b^k(\mathcal{N}^{(l)}) = b^k(\mathcal{N}), k = 1, 2, \dots, L-1\\
                & b^L(\mathcal{N}^{(l)}) = \mathbb{R}\\
            \end{align*}
        \item On weights:
            \begin{align*}
                & \omega^k(\mathcal{N}^{(l)}) = \omega^k(\mathcal{N}), k = 1, 2, \dots, L-2\\
                & \omega^{L-1}(\mathcal{N}^{(l)}) = \omega^{L-1}_l(\mathcal{N})\\
            \end{align*}
        \item On thresholds:
            \begin{align*}
                & \theta^k(\mathcal{N}^{(l)}) = \theta^k(\mathcal{N}), k = 1, 2, \dots, L-2\\
                & \theta^{L-1}(\mathcal{N}^{(l)}) = \theta^{L-1}_l(\mathcal{N})\\
            \end{align*}
        \end{enumerate}
\end{defn}

The Component Net operation is used to decompose a multi-labeled net into several nets, each can be viewed as a Characteristic MLP.

\begin{defn}[O-Product Net]
    Let $\mathcal{N}_1, \mathcal{N}_2$ be two L-layered MLPs with $b^1(\mathcal{N}_1) = b^1(\mathcal{N}_2)$ and $b^L(\mathcal{N}_1) = b^L(\mathcal{N}_2) = \mathbb{R}$, their \textbf{O-Product Net} is defined by the following equations:
    \begin{enumerate}
        \item On layer structure:
            \begin{align*}
                 & b^1(\mathcal{N}_1 \otimes \mathcal{N}_2) = b^1(\mathcal{N}_1), b^1(\mathcal{N}_1 \otimes \mathcal{N}_2) = b^1(\mathcal{N}_2)\\
                 & b^k(\mathcal{N}_1 \otimes \mathcal{N}_2) = b^k(\mathcal{N}_1) \times b^k(\mathcal{N}_2), k = 2, \dots, L-1\\
                 & b^L(\mathcal{N}_1 \otimes \mathcal{N}_2) = b^L(\mathcal{N}_1) \times b^L(\mathcal{N}_2) = \mathbb{R}^2\\
            \end{align*}
        \item On weights:
            \begin{align*}
                 & \omega^1(\mathcal{N}_1 \otimes \mathcal{N}_2) = 
            \left(\begin{array}{c}\omega^1(\mathcal{N}_1)\\\omega^1(\mathcal{N}_2)\end{array}\right)\\
                 &\omega^k(\mathcal{N}_1 \otimes \mathcal{N}_2) = diag(\omega^k(\mathcal{N}_1),\omega^k(\mathcal{N}_2)), k = 2, \dots, L-1\\
            \end{align*}
        \item On thresholds:
            \begin{align*}
                 &\theta^k(\mathcal{N}_1 \otimes \mathcal{N}_2) = 
            \left(\begin{array}{c}\theta^k(\mathcal{N}_1)\\\theta^k(\mathcal{N}_2)\end{array}\right) &k = 1, 2, \dots, L\\
            \end{align*}
        \end{enumerate}
\end{defn}

O-Product Net suits our needs when we attempt to integrate several Characteristic MLPs indicating different labels into one new net.

The next operation is for the scenarios when we want to calculate the sum, i-product or o-product of two nets with different layer structure. 
We can simply take the net with the smaller layer number, extend it several times until it has the same number of layers with the other, and carry out the desired operation.
The extension should maintain all the output values.
To accomplish this goal we have to replace the sigmoid function with the ReLU function in the last layer.
ReLU function is defined as
\[ReLU(x) = max(0, x)\]

\begin{defn}[Identical Extension-$\beta$]
    $\mathcal{N}$ is an L-layered MLP, $b^L(\mathcal{N}) = \mathbb{R}^{n_L}$, 
    its \textbf{Identical Extension} $T(\mathcal{N})$ is an $(L + 1)$-layered MLP defined by the following equations:
    \begin{enumerate}
        \item On layer structure:
            \begin{align*}
                & b^k(T(\mathcal{N})) = b^k(\mathcal{N}), k = 1, 2, \dots, L\\
                & b^{L+1}(T(\mathcal{N})) = b^{L}(T(\mathcal{N}))\\
            \end{align*}
        \item On weights:
            \begin{align*}
                & \omega^k(T(\mathcal{N})) = \omega^k(\mathcal{N}), \,\, k = 1, 2, \dots, L-1\\
                & \omega^{L}(T(\mathcal{N})) = \mathcal{I},\\
            \end{align*}
        \item On thresholds:
            \begin{align*}
                & \theta^k(T(\mathcal{N})) = \theta^k(\mathcal{N}), \,\, k = 1, 2, \dots, L-1\\
                & \theta^{L}(T(\mathcal{N})) = 0
             \end{align*}  
     \end{enumerate}
\end{defn}

\section{Summary}
As the author has explained in the introduction section, the central idea of this theory is that complex neural nets can be built from simpler ones. 
The building process require a good decomposition of the original dataset, and the results of the decomposition should be geometrically simple.
Different MLP algebra operations work with different kinds of decomposition, here is some naive decompositions:
\begin{enumerate}
        \item High dimensional data can be projected to a certain subspace, and I-Product Net operation can work with disjoint projections;
        \item Sum Net operation can work with dataset which have several connected components; 
        \item Datasets with several labels can be decomposed into several single-labeled subsets, and O-Product Net operation is designed for this kind of scenarios.
\end{enumerate}

One constaint of this theory is that it only works with MLP networks, but modern neural networks have more complex structures, such as feature maps and pooling layers in Convolutional Neural Networks(CNN) and circuits in Recurrent Neural Networks(RNN). 
The connection between MLP algebra and these networks is worth further exploration.
\bibliographystyle{hplain}
\bibliography{nn}
\nocite{Cybenko1989}
\nocite{Goodfellow-et-al-2016}
\nocite{Nielsen2015}
\nocite{DB2013}
\nocite{GKS}
\nocite{LR}
\end{document}